\documentclass[draftnocls,onecolumn,12pt]{IEEEtran}
\usepackage{amssymb}
\usepackage{amsmath}
\usepackage{stfloats}
\usepackage{graphicx}
\usepackage{subfigure}
\usepackage{tabularx}
\usepackage{epsfig,epsf,color,balance,cite}
\usepackage{multirow}
\usepackage{url}
\usepackage{amsthm}

\begin{document}
%
% paper title
\title{A Machine Learning Framework for Resource Allocation Assisted by Cloud Computing}

\author{Jun-Bo Wang, Junyuan Wang, Yongpeng Wu, Jin-Yuan Wang, \\Huiling Zhu, Min Lin, Jiangzhou Wang
\thanks{Jun-Bo Wang is with National Mobile Communications Research Laboratory, Southeast University, Nanjing 210096, China (email: jbwang@seu.edu.cn). }
\thanks{Junyuan Wang, Huiling Zhu and Jiangzhou Wang are with the School of Engineering and Digital Arts, University of Kent, Canterbury, Kent, CT2 7NT, United Kingdom (email:  \{jw712,h.zhu,j.z.wang\}@kent.ac.uk). }
\thanks{Yongpeng Wu is with the Department of Electronic Engineering, Shanghai Jiao Tong University, China (email: yongpeng.wu@sjtu.edu.cn)}
\thanks{Min Lin and Jin-Yuan Wang are with College of Telecommunications and Information Engineering, Nanjing University of Posts and Telecommunications, Nanjing 210003, China (email: \{linmin,jywang\}@njupt.edu.cn). }
\thanks{Corresponding author: Jun-Bo Wang\hfill Corresponding email: jbwang@seu.edu.cn }
}

\markboth{}
{}

% make the title area
\maketitle %\linespread{1.7}
\begin{abstract}
Conventionally, the resource allocation is formulated as an optimization problem and solved online with instantaneous scenario information. Since most resource allocation problems are not convex, the optimal solutions are very difficult to be obtained in real time. Lagrangian relaxation or greedy methods are then often employed, which results in performance loss. Therefore, the conventional methods of resource allocation are facing great challenges to meet the ever-increasing QoS requirements of users with scarce radio resource. Assisted by cloud computing, a huge amount of historical data on scenarios can be collected for extracting similarities among scenarios using machine learning. Moreover, optimal or near-optimal solutions of historical scenarios can be searched offline and stored in advance. When the measured data of current scenario arrives, the current scenario is compared with historical scenarios to find the most similar one. Then, the optimal or near-optimal solution in the most similar historical scenario is adopted to allocate the radio resources for the current scenario. To facilitate the application of new design philosophy, a machine learning framework is proposed for resource allocation assisted by cloud computing. An example of beam allocation in multi-user massive multiple-input-multiple-output (MIMO) systems shows that the proposed machine-learning based resource allocation outperforms conventional methods.

\end{abstract}


\IEEEpeerreviewmaketitle

\newpage
\baselineskip=10mm

\section{Introduction}
\label{section1}
With the rapid development of electronic devices and mobile computing techniques, worldwide societal trends have demonstrated unprecedented changes in the way wireless communications are used. It is predicted that the monthly traffic of smartphones around the world will be about 50 exabytes in 2021~\cite{BIB01}, which is about 12 times of that in 2016. Obviously, wireless communications have become indispensable to our society and involved many aspects of our life. Many familiar scenarios such as ultra-dense residential areas and office towers, subways, highways, and high-speed railways challenge the future mobile networks in terms of ultra-high traffic volume density, ultra-high connection density, or ultra-high mobility. Due to its ability to guarantee the users' Quality of Service (QoS) and optimize the usage of facilities to maximum operators' revenue, how to allocate radio resources more efficiently is always one hot topic for future wireless communications~\cite{CSGK17}.

%To tackle these challenges, the cloud radio access network (CRAN) has been proposed as a combination of emerging technologies from both the wireless and computing technology industries by incorporating cloud computing into radio access networks (RANs)~\cite{BIB04}. Compared with traditional cellular architecture, CRAN are expected to supply lower operation and maintenance costs, higher capacity and spectral efficiency, smarter adaptability to non-uniform traffic, and more flexible network extensibility. Due to the ability to support all existing and future mobile communication standards, CRAN has been considered as a promising future mobile network architecture. Several specific architectures with different functional splits have been proposed by the academia and industry, such as heterogeneous CRAN (H-CRAN), Fog-RAN (F-RAN), and
%RAN-as-a-Service (RANaaS).

%To tackle these challenges, many key enabling technologies have been identified.

In practical networks, the overall performance depends on how to exploit the fluctuation of wireless channels and traffic loads to efficiently and dynamically manage the hyper-dimensional radio resources (such as frequency bands, time slots, orthogonal codes, transmit power, and transmit-receive beams) and fairly to support users' QoS requirements. On one hand, radio resources are inherently scarce, since all users competitively share the common electromagnetic spectrum and wireless infrastructures. On the other hand, wireless services have been becoming increasingly sophisticated and various, each of which has a wide range of QoS requirements. Efficient and robust resource allocation algorithms are essential for the success of future mobile networks. Conventionally, the resource allocation problems are often formulated mathematically as optimization problems. After collecting instantaneous channel state information (CSI) and QoS requirements of users, the formulated optimization problems are solved online. That is, the solutions must be obtained shortly since wireless channels and traffic loads are varying quickly. However, most of the optimization problems are not convex~\cite{BIB09}, which indicates that the optimal solutions are often very difficult to be obtained, especially in the scenarios with a lot of users and diverse radio resources. Therefore, conventional Lagrangian relaxation or greedy methods are often employed to find solutions online. Inevitably, the online solutions of resource allocation will result in performance loss. With the increasing of users' QoS requirements, conventional methods are facing great challenges in designing more sophisticated resource allocation schemes to further improve system performance with scarce radio resource, which motivates the exploration of novel design philosophy for resource allocation.

%In March 2016, a five-game Go match was held between 18-time world champion Lee Sedol and AlphaGo, a computer Go program developed by Google DeepMind~\cite{BIB05}. It is estimated that the number of possible Go games is $ 2.082\times10^{170}$ (i.e., a 2 followed by 170 zeroes) which is much larger than the number of atoms in the universe (around $10^{80}$). From the views of conventional computing theory, Go had previously been regarded as an extremely difficult problem that was expected to be out of reach for the technology of the time. Guided by machine learning, AlphaGo finds its moves based on the knowledge previously ``learned" from historical match records. Surprisingly, AlphaGo won all but the fourth game. It is the first time that artificial intelligence (AI) beats a top human professional player at the game on a full-sized board, which was also recognized as a major milestone of machine learning research.

%\textcolor[rgb]{1.00,0.00,0.00}{As a new design method, machine learning has drawn much attention recently \cite{7932863,7935536}. }
In March 2016, a five-game Go match was held between 18-time world champion Lee Sedol and AlphaGo, a computer Go program developed by Google DeepMind~\cite{BIB05}. From the views of conventional computing theory, Go had previously been regarded as an extremely difficult problem that was expected to be out of reach for the state-of-the-art technologies. Surprisingly, AlphaGo found its moves based on the knowledge previously ``learned" from historical match records and won all but the fourth game. {Inspired by the victory of AlphaGo, how to apply machine learning techniques to address the challenges in future communications attracts great attention and has been discussed widely~\cite{7932863,7935536}.}

In practical wireless communications, the radio resources are dynamically allocated according to the instantaneous information including CSI and QoS requirements of users. Inexpensive cloud storage makes it very easy to save the information as data on historical scenarios that previously we would have ignored and trashed. Recent investigations have found that these data convey a lot of similarities between current and historical scenarios on user requirements and wireless propagation environments~\cite{BIB07}. Using the similarities among scenarios, the solutions of resource allocation in historical scenarios can be exploited to improve the resource allocation of current scenario. More specifically, the solutions of resource allocation in historical scenarios can be searched offline and stored in advance. When the measured data of current scenario arrives, it is not necessary to use conventional Lagrangian relaxation or greed methods to solve the resource allocation problem online. Instead, we only need to compare the current scenario with historical scenarios and find the most similar one. Then, we use the solution of the most similar historical scenario to allocate the radio resources for the current scenario. Interestingly, the offline characteristic makes it possible to use advanced cloud computing techniques to find optimal or near-optimal solutions of resource allocation for historical scenarios, which can improve the performance of resource allocation accordingly.

\section{Mathematical Modeling of Resource Allocation}

\label{section2}
\begin{figure}[!t]
\centering
\includegraphics[width=0.6\textwidth]{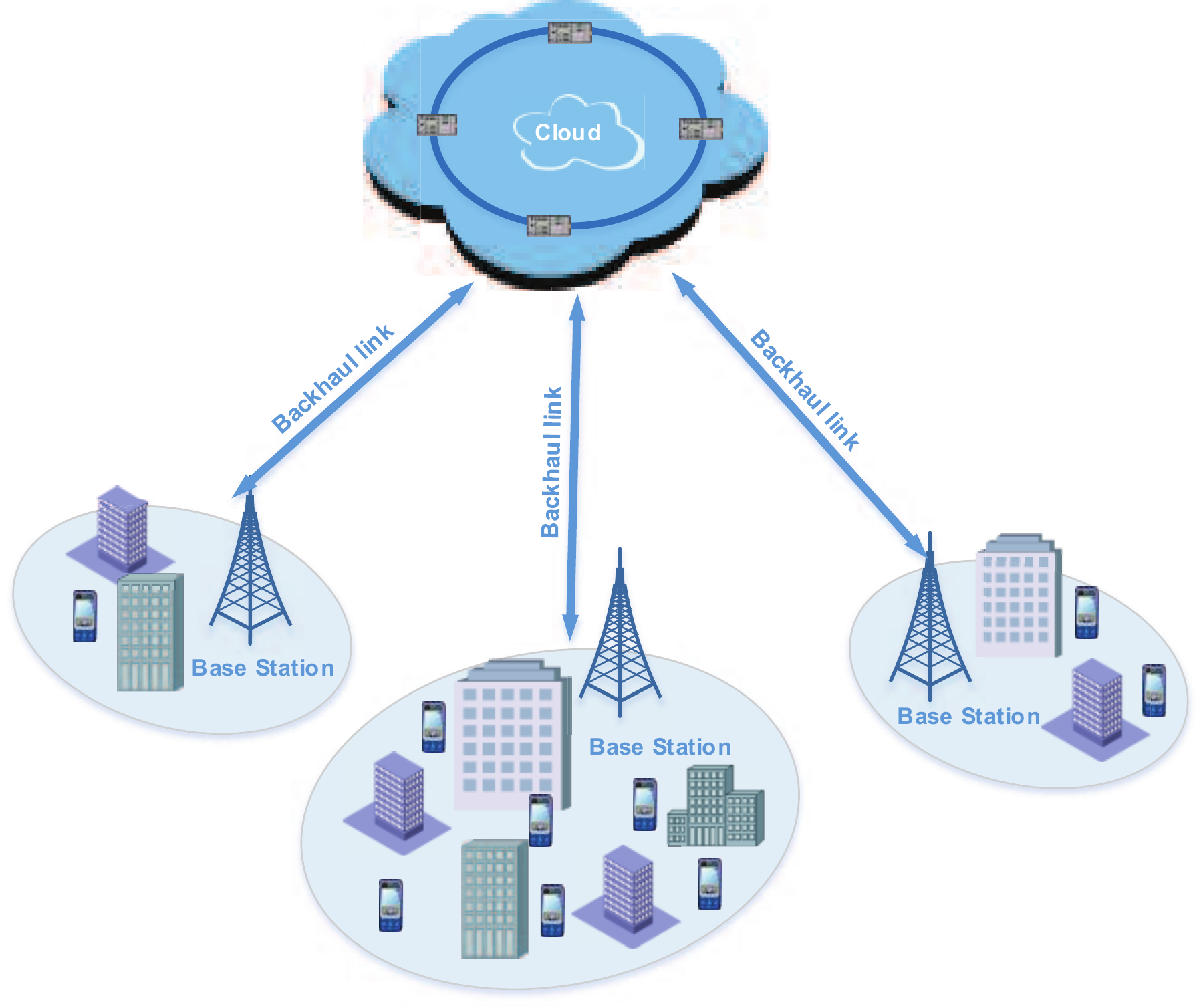}
\caption{Wireless Communications Assisted by Cloud Computing~\label{fig1}}
\end{figure}

As illustrated in Fig.~\ref{fig1},
the architecture of wireless communications assisted by cloud computing consists of three main components,
(i) configurable computing resources clustered as a cloud with high computational and storage capabilities,
(ii) base station (BS) with wireless access functions, and
(iii) backhaul links {which deliver the measured data of real scenarios from the BS to the cloud and deploy the machine learning based resource allocation schemes at the BS. More details will be discussed in the next section. In general, the resource allocation which is preformed at the BS} can be formulated as a mathematical optimization problem~\cite{BIB09}, given by
\begin{eqnarray}
\label{equ1}\mathop {{\mathop{\rm miniminze}\nolimits} }\limits_{{\bf{x}} \in S}  && f\left( {{\bf{x}},{\bf{a}}} \right)\\
{\rm{subject}}\;{\rm{to}}&& {g_i}\left( {{\bf{x}},{\bf{a}}} \right) \le {\rm{0}}\quad i = 1, \cdots ,m\nonumber\\
&&{h_i}\left( {{\bf{x}},{\bf{a}}} \right) = {\rm{0}}\quad i = 1, \cdots ,p\nonumber
\end{eqnarray}
where $\bf{x}$ is the variable vector of the problem,
$f(\cdot,\cdot)$ is the objective function to be minimized over the vector $\bf{x}$,
$\bf{a}$ is the parameter vector that specifies the problem instance,
$\left\{ {{g_i}} \right\}_{i = 1}^m$ and $\left\{ {{h_i}} \right\}_{i = 1}^m$ are called inequality and equality constraint functions, respectively, and $S$ is called a constraint set.
By convention, the standard form defines a minimization problem.
A maximization problem can be treated by negating the objective function.

If a resource allocation problem is formulated as the form~(\ref{equ1}), all elements in the vector $\bf{x}$ are referred as variables which describe the allocated amount or configuration of radio resources, such as the transmit power level, and the assigned subcarrier index. All elements in the vector $\bf{a}$  are the system parameters or wireless propagation parameters, such as the bandwidth, the subcarrier number, and the background noise level. $\left\{ {{g_i}} \right\}_{i = 1}^m$ and $\left\{ {{h_i}} \right\}_{i = 1}^m$ are used to define the specific scenario and the limitations on the resource allocation, such as the available amount of radio resources, users¡¯ QoS requirements, and the impacts from all kinds of interferences and noises. The objective function describes the characteristics of the best possible solution and reveals the design objective, i.e., the key performance metrics for resource allocation. {For a specified scenario described by $\bf{a}$, the optimal solution of resource allocation $\bf{x}^{*}$ is the vector that obtains the best value of objecitve function among all possible vectors and satisfies all constraints.}

%\section{Overview of Wireless Communications Assisted by Cloud Computing }
\section{A Machine Learning Framework}

For existing wireless systems assisted by cloud computing,
a huge amount of data on historical scenarios may have been collected and stored at the cloud.
{The strong computing capability of the cloud is exploited to search the optimal or near-optimal solutions for these historical scenarios.
By classifying these solutions, the similarities hidden in these historical scenarios are extracted as a machine learning based resource allocation scheme.
The machine learning based resource allocation scheme} will be forwarded to guide BS how to allocate radio resource more efficiently. When a BS is deployed in a new area, there is usually no available data about historical scenarios. In this case, the initially historical data can be generated from an abstract mathematical model with realistic BS locations, accurate building footprints, presumptive user distribution and requirements, and wireless propagation models. When the new BS emerges into service, the measured data of real-time scenarios will be collected from practical systems, and later used as historical data for learning.

\begin{figure}[!t]
{\centering
\includegraphics[width=1\textwidth]{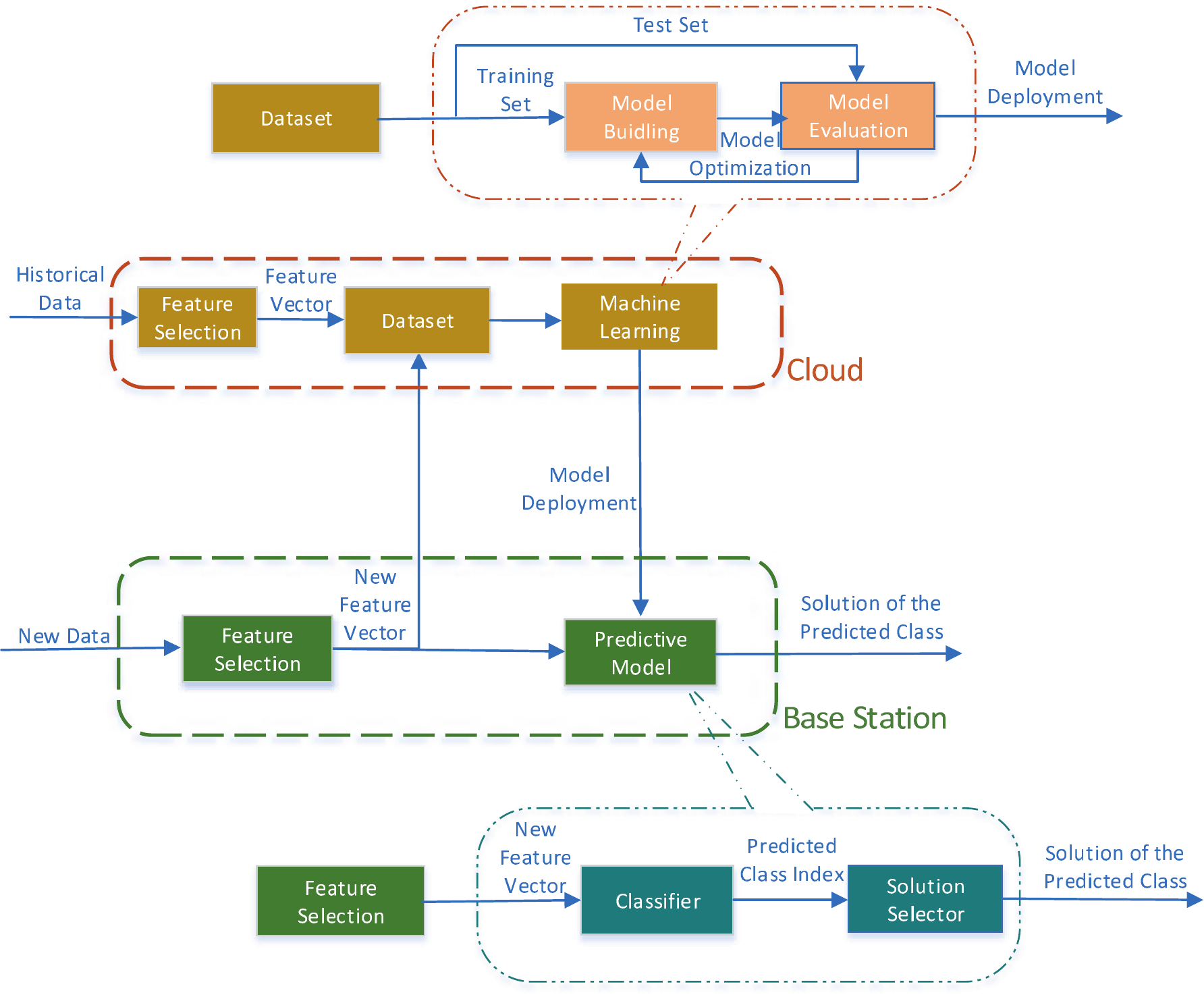}
\caption{A Machine Learning Framework of Resource Allocation. \label{fig2}}}
\end{figure}

%When a CRAN is deployed in a new area, there is usually no available raw data about historical wireless accesses. In this case, the initial raw data can be generated from an abstract mathematical model with realistic RRH locations, accurate building footprints, presumptive user distribution and requirements, and wireless propagation models. When the new CRAN comes into service, the measured data of real-time scenario will be collected from practical systems, and later used as historical raw data for learning. For an existing CRAN which needs to update the function of resource allocation, a huge amount of historical raw data may have been stored and available.

The proposed machine learning framework is shown in Fig.~\ref{fig2}.
At the cloud, a huge amount of historical data on scenarios are stored using the cloud storage.
The historical data has a lot of attributes, including the user number, the CSI of users, international mobile subscriber identification numbers (IMSIs) of users, and so on. Some attributes, such as IMSIs of users, may be irrelevant for the specific resource allocation, i.e., these irrelevant attributes are not included in the parameter vector $\bf{a}$ in the optimization problem~(\ref{equ1}). Learning from a large number of raw data with many attributes generally requires a large amount of memory and computation power, and it may influence the learning accuracy~\cite{BIB10}. Therefore, the irrelevant attributes can be removed without incurring much loss of the data quality. In order to reduce the dimensionality of the data and enables the learning process to operate faster and more effectively, feature selection is carried out to identify and remove as many irrelevant attributes as possible, which will be discussed in Section~\ref{FeatureSelection}.

Through feature selection, some key attributes are selected from the historical data and presented as a feature vector. However, there may exist some operation faults in the data measurement, transmission, and storage, which results in the abnormal, incomplete or duplicate values in feature vectors. Therefore, necessary preprocessing is required to delete erroneous or duplicate feature vectors. Then, all remain feature vectors are collected to form a very large dataset. Further, all feature vectors in dataset are split randomly into a training and a test set. Normally, 70-90\% of the feature vectors is assigned into the training set.

With the training set, a supervised learning algorithm in machine learning is adopted to find the similarities hidden in historical data. By doing so, a predictive model can be built which will be used to make resource allocation decision for future unexpected scenario. More specifically, with the aid of cloud computing, advanced computing techniques can be used to search the solutions for the optimization problem~(\ref{equ1}) with more computational time. Compared with conventional Lagrangian relaxation or greedy methods, the performance of searched solutions can be improved significantly. Therefore, a high performance solution of resource allocation can be searched offline and associated with each training feature vector, which will be discussed in Section~\ref{SolutionofProblem}. All training feature vectors with the same solutions are classified into one class and each class is associated with its own solution. The resource allocation problem is now transformed into a multiclass classification problem, which will be discussed in Section~\ref{MultiClassProblem}. In order to solve the multiclass classification problem, a predictive model will be built with two functions. The first is to predict the class for future scenario, which can be mathematically described as a classifier $l=Classifier(F_T)$. $F_T$ is the input feature vector extracted from scenario, and $l$ is the output class index showing that the scenario belongs to the $l^{\text{th}}$ class. Then, the associated solution of the $l^{\text{th}}$ class is selected to allocate radio resources for the scenario depicted by $F_T$. Before deploying the model, the recently built predictive model is evaluated by the test set and further optimized until the evaluation results are satisfactory.

Using the backhaul links, the built predictive model and the associated solutions of all classes will be transmitted to BS. At the BS, the measured data of a real-time scenario is first used to form its new feature vector. Then the new feature vector will be input into the the built predictive model to allocate radio resource. Meanwhile, the new feature vector will be collected and stored temporarily at BS and forwarded to the cloud later for updating the dataset, which is very important for tracing the evolutions of real scenarios, including user behaviors and wireless propagation environments.

Although a lot of computing resources are consumed to build a predictive model, the computing work can be carried out offline during the off-peak time. Moreover, the dataset updating and the model deployment can also be accomplished during the off-peak time. Therefore, the cloud can be shared with multiple BSs and the computing tasks can be flexibly scheduled to make full use of the available computing resources.

\section{Application of Supervised Learning to Resource Allocation}
In the proposed machine learning framework, a machine learning algorithm is adopted to build a predictive model.
General speaking, machine learning algorithm is usually categorized as either supervised or unsupervised~\cite{BIB07}. In the supervised learning, the goal is to learn from training data which are labeled with nonnegative integers or \emph{classes}, in order to later predict the correct response when dealing with new data. The supervised approach is indeed similar to human learning under the supervision of a teacher. The teacher provides good training examples for the student, and the student then derives general rules from these specific examples. In contrast to supervised learning, the data for unsupervised learning have no labels and the goal instead is to organize the data and find hidden structures in unlabeled data. Most machine learning algorithms are supervised. In the following, we will discuss how to apply the supervised learning to solve the resource allocation problem.

\subsection{Feature Selection}
\label{FeatureSelection}

%The raw data has a lot of attributes which are either irrelevant or redundant. Learning from a large number of raw data with many attributes generally requires a large amount of memory and computation power, also it may influence the learning accuracy~\cite{BIB10}. Therefore, the irrelevant or redundant attributes can be removed without incurring much loss of the data quality. Feature extraction is the process of identifying and removing as many irrelevant and redundant attributes as possible, which reduces the dimensionality of the data and enables machine learning algorithms to operate faster and more effectively.

In machine learning, feature selection, also known as attribute selection, is the process of selecting a subset of relevant attributes in historical data to form feature vector for building predictive models. The selection of an appropriate feature vector is critical due to the phenomenon known as ``the curse of dimensionality"~\cite{Curese}. That is, each dimension that is added to the feature vector requires exponentially increasing data in the training set, which usually results in practical significant performance degradation. Therefore, it is necessary to find a low dimension of feature vectors that captures the essence of resource allocation in practical scenarios.

In order to reduce the dimensionality of feature vectors, only valuable information for the resource allocation can be selected as features. After modeling the resource allocation as the optimization problem~(\ref{equ1}), all valuable information is included in the parameter vector $\bf{a}$. Observing the elements of $\bf{a}$, it can be found that they can be further divided into two categories: time-variant (dynamic) or time-invariant (static). Some elements are constants and thus labeled as time-invariant parameters, such as subcarrier number, maximum transmit power, and antenna number. Other elements that change quickly and are required to be measured and feedback all the time for making decisions of the resource allocation are labeled as time-variant parameters, such as user number, CSI of all users, and interference levels. As the time-invariant parameters keep unchanged, in order to minimize the dimension of the feature vectors, only the time-variant parameters can be considered to be features. Moreover, some time-variant parameters cannot be selected as features since it may be redundant in the presence of another relevant feature with which it is strongly correlated. In short, an individual feature vector specifies a unique scenario for resource allocation. However, it should be noted that the feature selection is a process of trial and error, which can be time consuming and costly especially with very large datasets.

\subsection{Solutions of Optimization Problems}
\label{SolutionofProblem}

To facilitate the application of supervised learning, the solution of resource allocation problem specified by each training feature vector should be obtained in advance. Then, each training feature vector is associated with its solution. According to the associated solutions, all feature vectors are labeled into multiple classes. More specifically, all training feature vectors with the same solution are placed with the same class label, indexed by a nonnegative integer. In other words, each class is associated with its unique solution. The class label information of all training feature vectors will be used to build a predictive model. In practice, the measured data of real-time scenario is selected as a new feature vector. Then the predictive model will predict the class for the new feature vector, and output the associated solution of the predicted class, i.e., how to allocate the radio resource for the real-time scenario. Obviously, if too many training feature vectors are associated with low performance solutions, the built predictive model cannot supply high performance solutions for practical resource allocation. Therefore, finding optimal or near-optimal solutions of all training feature vectors is crucial for building a high performance predictive model.

In the resource allocation problem~(\ref{equ1}), all elements in the vector $\bf{x}$ are used to describe how to allocate the radio resources. Mathematically, the allocation of many radio resources can be described by integer variables, such as subcarriers, timeslots, modulation and coding schemes. Intuitively, the transmit power level can be adjusted arbitrarily between the maximum transmit power and zero. It seems that that only a continuous variable can be used to describe the transmit power allocation. However, in order to simplify the system complexity, the transmitter in practical systems are usually allowed to transmit signals with only a few prefixed power levels. Therefore, most practical resource allocation issues can be modeled as an integer optimization problem. When the number of integer variables in an integer optimization problem is very small, the optimal solution can be found by exhaustive search. However, if there are many integer variables, finding an optimal solution of resource allocation is extremely computationally complex because they are known to be non-deterministic polynomial-time hard (NP-hard)~\cite{Combination}. In this case, it is more feasible to search the near-optimal solutions for all training feature vectors. Moreover, the offline characteristic of building model and the strong cloud computing and storage capabilities make it possible to spend more computation time using the metaheuristics to search near-optimal solutions. Some famous metaheuristics algorithms~\cite{BIB12}, such as particle swam optimization (PSO) and ant colony optimization (ACO) have been applied to the solution of many classic combination optimization problems. For most of these applications, the results show that these metaheuristics algorithms outperform other algorithms, including conventional Lagrangian relaxation or greed based algorithms.

\subsection{Multiclass Classification Problem and Classifier}
\label{MultiClassProblem}
\begin{figure}[!t]
\centering
\includegraphics[width=0.50\textwidth]{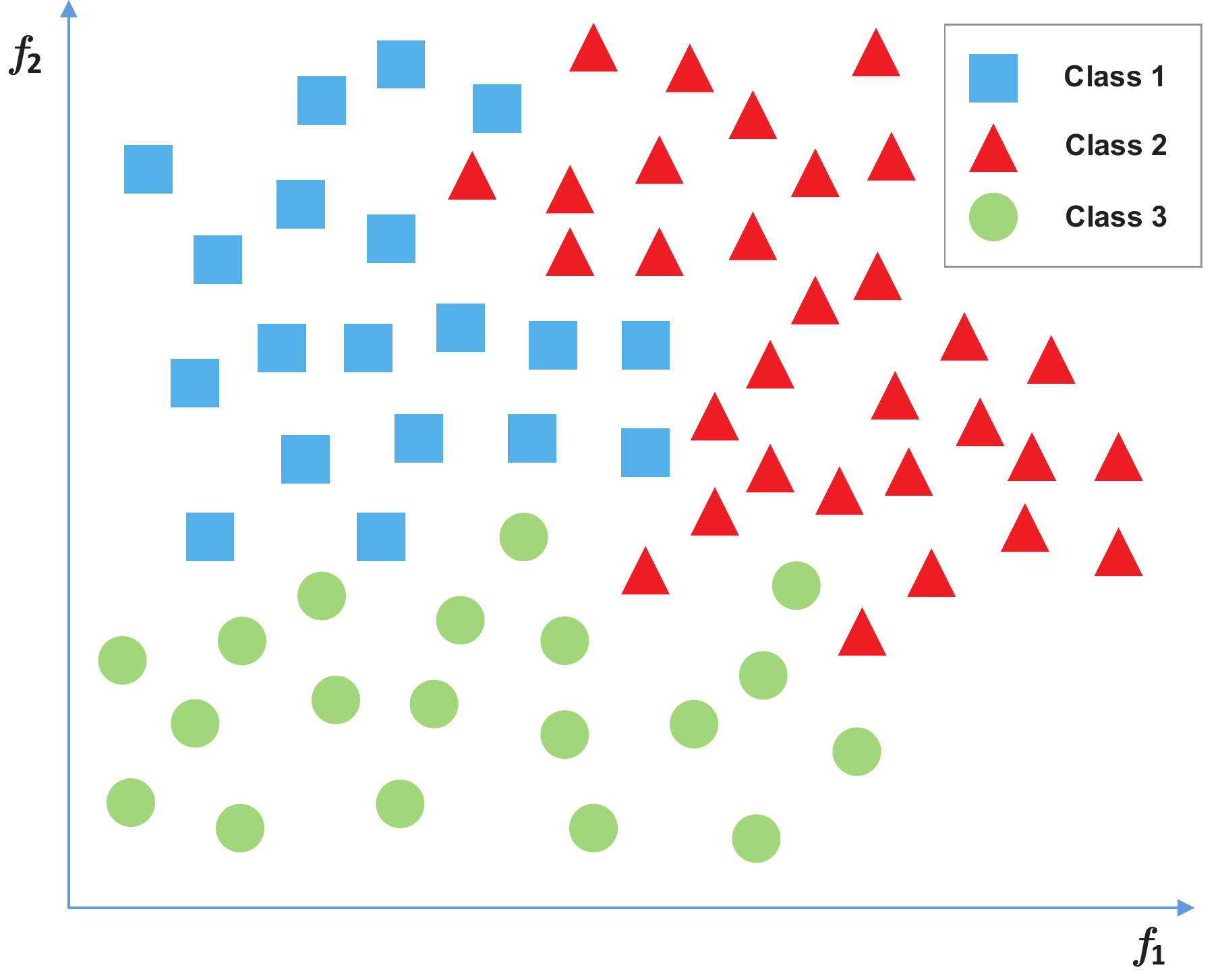}
\caption{Multiclass classification \label{fig3}}
\end{figure}

When the class label information is ready for all training feature vectors, it starts to look for the similarities hidden in labeled feature vectors. Mathematically, the set of all possible feature vectors constitutes a feature space. As a special case, if the order of the feature vectors is 2, the feature space is a two-dimensional space, as shown Fig.~\ref{fig3}. Note that $f_1$ and $f_2$ are the first and second elements of feature vectors, respectively. When all labeled feature vectors are shown in the feature space, it can be observed that the feature vectors with the same class label are often distributed very closely. Accordingly, the feature space can be
divided into several subspaces and most feature vectors with the same class label are located within the same subspace. Then, the hidden similarities can be exploited by building a classifier
, which predicts the class of new feature vector by determining which subspace it is located in. In supervised learning, such learning process is often called as a multiclass classification problem~\cite{BIB13}. So far, many machine learning algorithms have been used for designing multiclass classifiers. Selecting a machine learning algorithm is also a process of trial and error. It is a trade-off between specific characteristics of the algorithms, such as computational complexity, speed of learning, memory usage, predictive accuracy on new feature vectors, and interpretability. Therefore, the design of multiclass classifier is an essential task.

\begin{figure}[!t]
\centering
\includegraphics[width=0.50\textwidth]{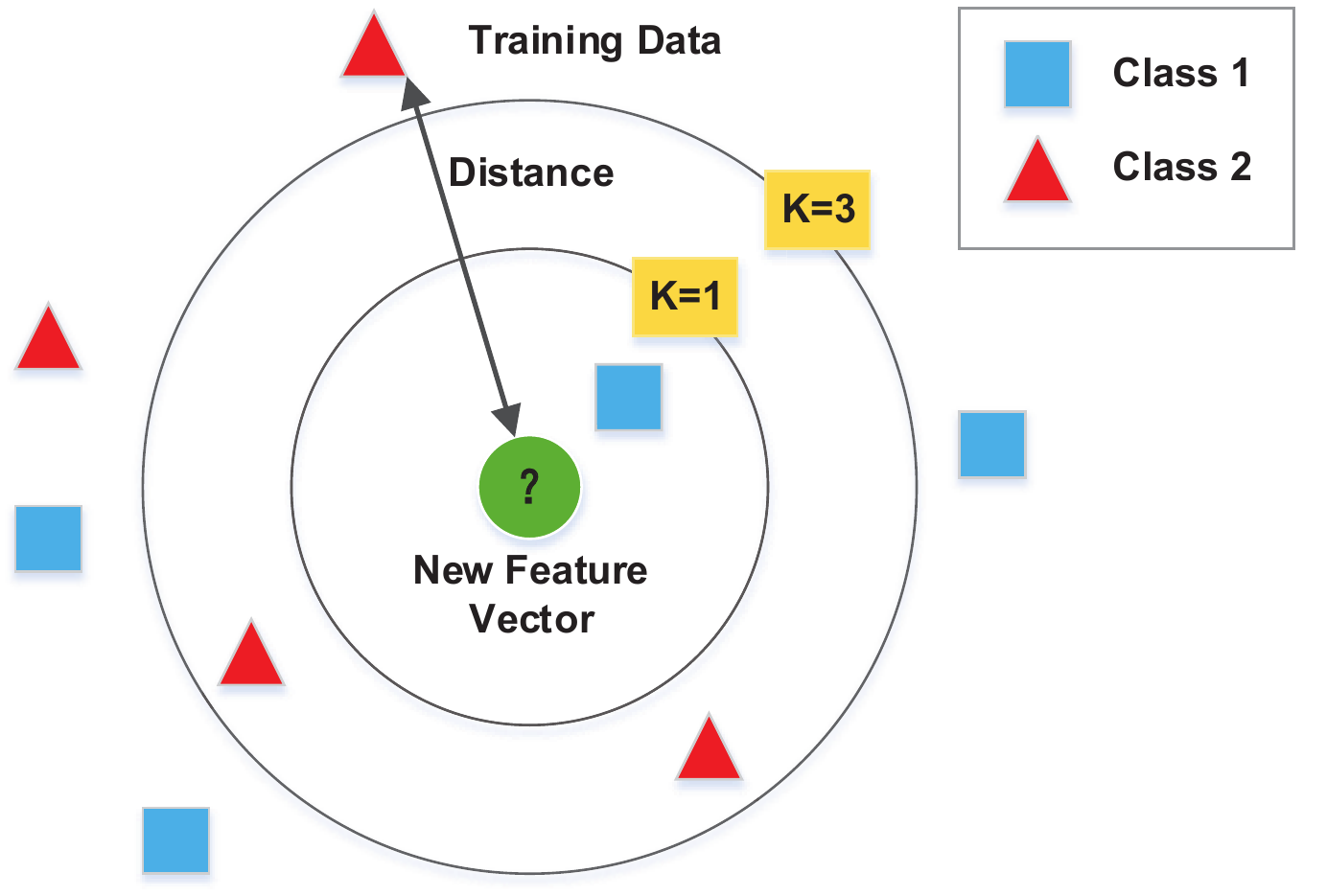}
\caption{$k$-NN algorithm \label{fig4}}
\end{figure}

Here, we briefly introduce the $k$-NN algorithm which is known to be very simple to understand but works incredibly well in practice. As shown in~Fig.~\ref{fig4}, whenever a new feature vector arrives, the $k$-NN algorithm picks up totally $k$ nearest neighbors of the new feature vector from the training set. Then, the new feature vector is judged to belong to the most common class among its $k$ nearest neighbors. If $k = 1$, the new feature vector is simply categorized to the class of its nearest neighbor.

\section{Example: Beam Allocation in Multiuser Massive MIMO}
\label{section4}

In this section, the beam allocation problem in a single-cell multiuser massive MIMO system considered in \cite{Junyuan_Beam} will be taken as an example to demonstrate the efficiency of our proposed machine learning framework of resource allocation.

\begin{figure}[t]
\centering
\includegraphics[width=0.50\textwidth]{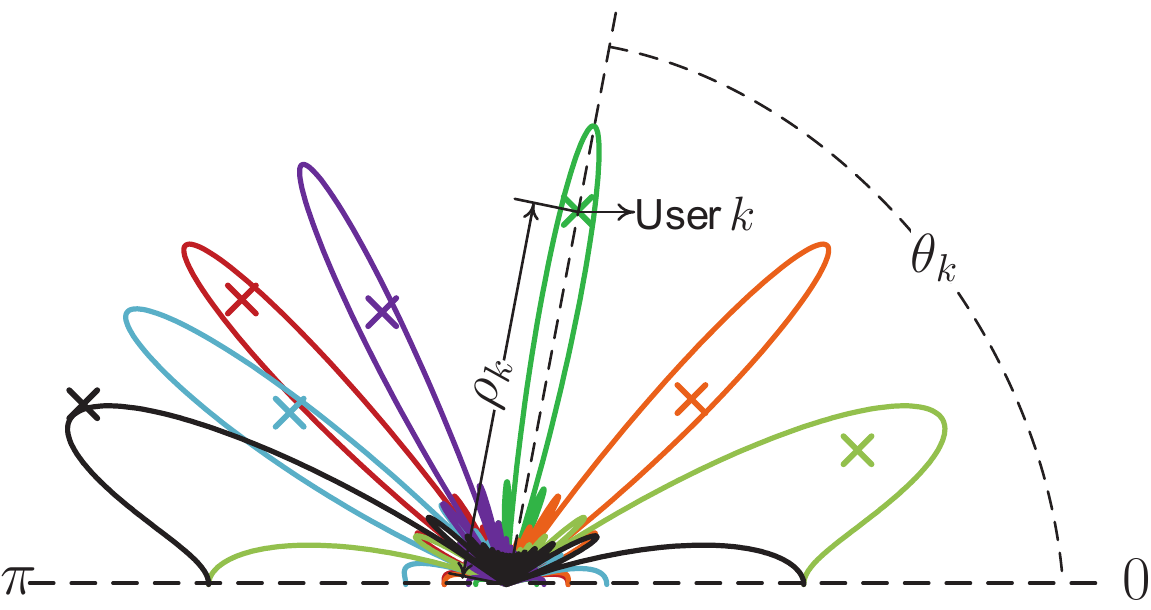}
\caption{Illustration of beam allocation in multiuser massive MIMO systems. ``x'' represents a user. A beam is allocated to the user in the same color. ($\rho_{k}, \theta_{k}$) denotes the polar coordinate of user $k$.}
\label{FIG_Beam}
\end{figure}

In the single-cell system, it is assumed that the BS is located at the center of the circular cell and $K$ users are uniformly distributed within the cell with unit radius, and each user is equipped with a single antenna. A massive mumber of $N\gg K$ fixed beams are formed by deploying the Butler network \cite{Butler} with a linear array of $N$ identical isotropic antenna elements at the BS. In such a fixed-beam system, a user will be served by a beam allocated to it as shown in Fig. \ref{FIG_Beam} where each user is served by the beam in the same color and $(\rho_{k}, \theta_{k})$ denotes the polar coordinate of user $k$. To serve multiple users simultaneously, the key problem is: how to efficiently allocate beams to users to maximize the sum rate?

As the number of beams $N$ is much larger than the number of users $K$ and each user is served by one beam, only some of beams will be active for serving users. Therefore, we first need to decide which beams are active. This can be solved by applying our machine learning framework. Specifically, the active beam solution serves as the output of the predictive model. By assuming a line-of-sight (LoS) channel, as the beam gains of $K$ users from $N$ beams are determined by the $K$ users' locations, the user layout $u=[(\rho_{1}, \theta_{1}), (\rho_{2}, \theta_{2}), \cdots, (\rho_{K}, \theta_{K})]$ should serve as the input data, which contains both the radial distance and phase information. Since the beam gains from various beams for a user significantly vary with its phase as shown in Fig. \ref{FIG_Beam}, the achievable sum rate with a beam allocation solution is mainly determined by the phase information of $K$ users. Therefore, the feature vector $F_{u}$ of a user layout data $u$ is selected as
\begin{align}\label{feature}
F_{T}=[\cos\theta^{(1)}, \cos\theta^{(2)}, \cdots, \cos\theta^{(K)}],
\end{align}
where $\theta^{(1)}\leq \theta^{(2)}\leq \cdots \leq \theta^{(K)}$ is the order statistics obtained by arranging $\theta_{1}, \theta_{2}, \cdots, \theta_{K}$.

\begin{figure*}[t]
\begin{center}
{\subfigure[$N=16$, $K=8$, $10^6$ training data.]
{\includegraphics[width=0.48\textwidth]{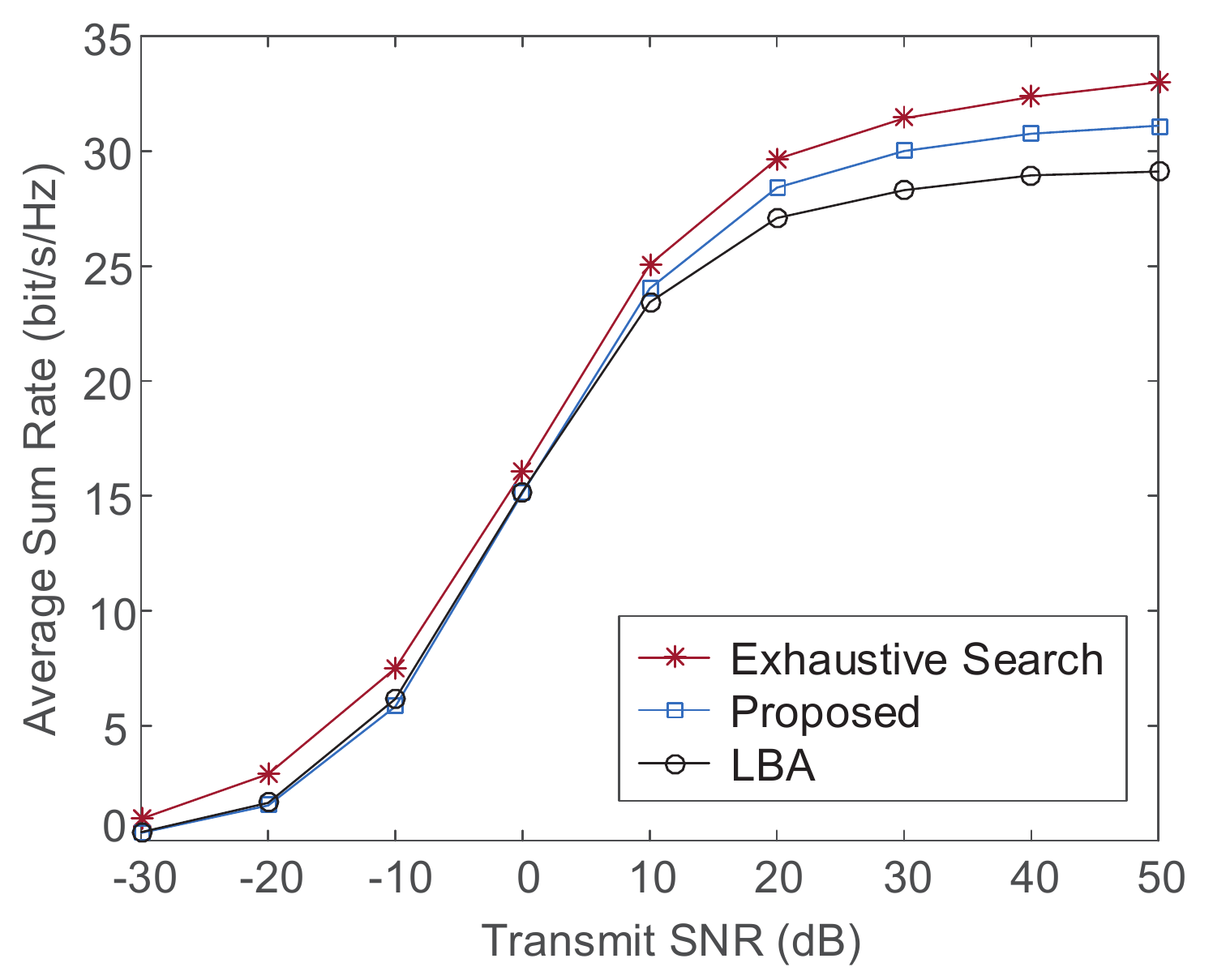}\label{FIG_AveR_SNR}} \hfil
\subfigure[$N=8$, transmit SNR=20dB.]
{\includegraphics[width=0.48\textwidth]{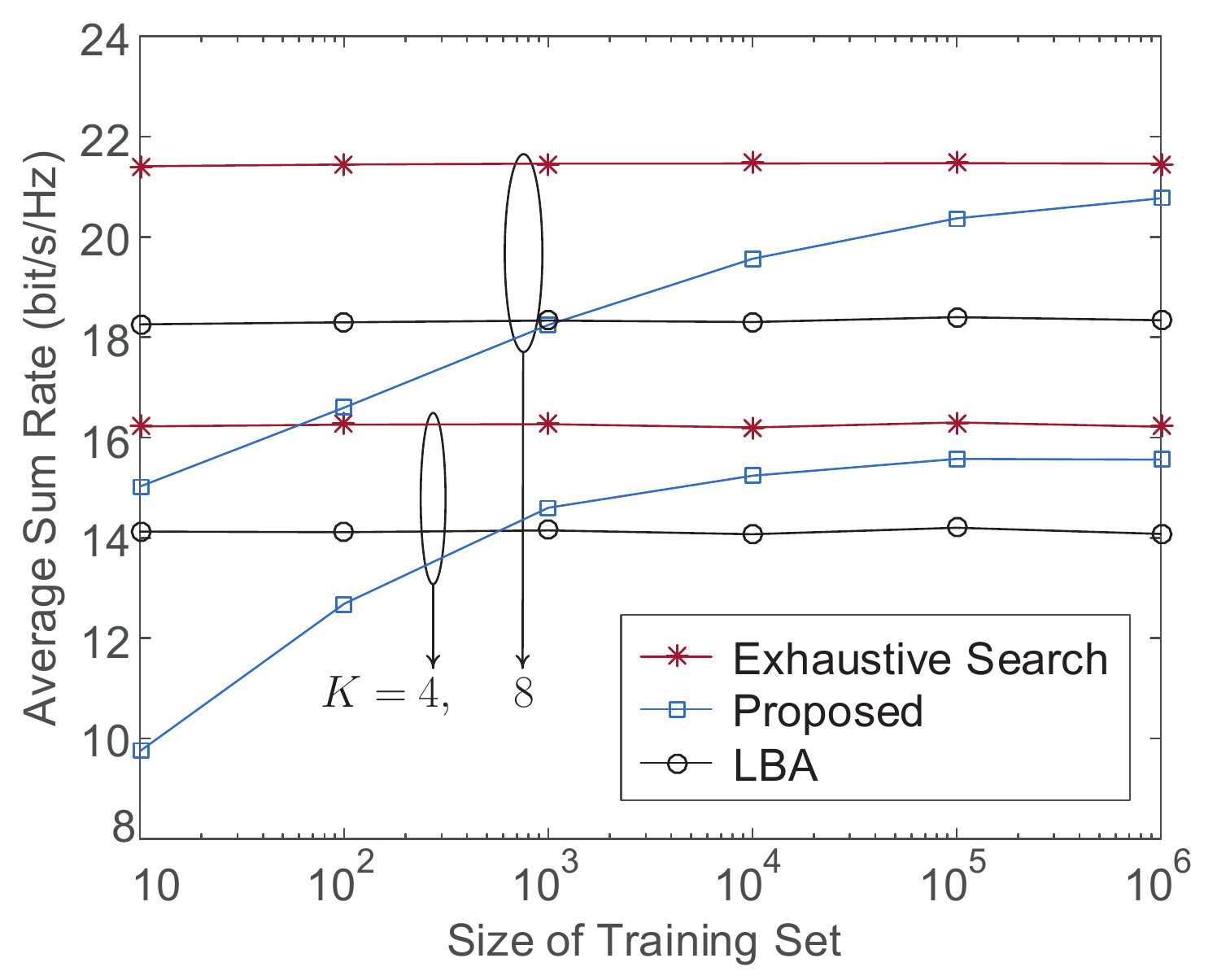}\label{FIG_AveR_S}}}
\caption{Average sum rate with our proposed machine learning framework, optimal exhaustive search and LBA algorithm proposed in \cite{Junyuan_Beam}. For the employed $k$-NN algorithm, $k=1$.}
\label{FIG_AveR}
\end{center}
\end{figure*}

Before performing resource allocation, we first need to train the predictive model by learning from a large amount of training user layout data, which can be generated by computer according to its distribution. For each training user layout, its feature vector formed according to (\ref{feature}) is associated with its active beam solution which can be obtained by employing offline beam allocation algorithms. Thanks to the strong cloud computing capability, optimal exhaustive search or near-optimal metaheuristics algorithms can be adopted as mentioned in Section~\ref{SolutionofProblem}. In this section, exhaustive search is applied for demonstration by assuming a smaller number of users and beams. After associating each feature vector in the training set with its active beam solution, all the training feature vectors are naturally classified into a variety of classes according to their active beam solutions. Specifically, the feature vectors sharing the same active beam solution are in the same class. A predictive model of active beam solution can be then built by applying a simple $k$-NN algorithm\footnote{The $k$-NN algorithm is employed in this section for illustration thanks to its simplicity.}, which can be then evaluated and optimized to guarantee its performance as Fig.~\ref{fig2} shows. For instance, it can be improved by adding more training data. The effect of the size of training set will be discussed by presenting Fig. \ref{FIG_AveR_S}.

The built predictive model is then deployed at the BS for beam allocation. For a new user layout $u_{i}$, by forming its feature vector $F_{u_{i}}$ and defining the distance from its feature vector $F_{u_{i}}$ to a stored training feature vector $F_{T_{j}}$, $d_{u_{i}, T_{j}}$, as
\begin{align}\label{distance}
d_{u_{i}, T_{j}}=\|F_{u_{i}}-F_{T_{j}}\|^2,
\end{align}
$k$ nearest neighbor feature vectors with $k$ smallest distances are picked. According to the $k$-NN algorithm, the most common class among these $k$ neighbors is chosen as the predictive class of the input user layout $u_{i}$ and the predictive model outputs the associated active beam solution of its predicted class. Based on the active beam information, each active beam is allocated to its best user with the highest received signal-to-interference-plus-noise ratio (SINR) by assuming equal power allocation among users. In addition, the new feature vectors $F_{u_{i}}$ will be collected to further update the dataset and trace the evolution of user layout.

%\begin{figure}[t]
%%\setcounter{totalnumber}{1}
%\centering
%\includegraphics[width=0.5\textwidth]{Beam.pdf}
%\caption{Illustration of beam allocation in multiuser massive MIMO systems. ``x'' represents a user. A beam is allocated to the user in the same color. ($\rho_{k}, \theta_{k}$) denotes the polar coordinate of user $k$.}
%\label{FIG_Beam}
%\end{figure}

Fig. \ref{FIG_AveR} presents the average sum rate with our proposed machine learning framework of beam allocation versus the transmit signal-to-noise ratio (SNR) and size of training set. For comparison, the average sum rate with both optimal exhaustive search and low-complexity beam allocation (LBA) proposed in \cite{Junyuan_Beam} are also plotted. It can be seen from Fig. \ref{FIG_AveR_S} that as the number of training data increases, the average sum rate achieved by our proposed machine learning framework increases and gradually approaches that with the optimal exhaustive search. It can also be observed from Fig. \ref{FIG_AveR} that with a larger training set, our algorithm outperforms the LBA algorithm proposed in \cite{Junyuan_Beam}, indicating that our proposed machine learning framework of resource allocation outperforms conventional techniques.

Note that for the aforementioned $k$-NN algorithm, the distances between new data and existing training data are calculated in real time. As a result, with a large number of training data, the computation complexity would become very high in practical systems. It is therefore important to design a low-complexity multiclass classifier, which will be discussed in Section~\ref{LowComplexity}.
%Fig. \ref{FIG_AveR} presents the average sum data rate with our proposed data-driven beam allocation algorithm versus the transmit signal-to-noise ratio (SNR) and size of training dataset. For comparison, the average sum data rate with both optimal exhaustive search and low-complexity beam allocation (LBA) proposed in \cite{Junyuan_Beam} are also plotted. It can be seen from Fig. \ref{FIG_AveR_S} that as the number of training samples increases, the average sum data rate achieved by our proposed data-driven framework increases and gradually approaches that with optimal exhaustive search. It can also be observed from Fig. \ref{FIG_AveR} that with a larger training dataset, our algorithm outperforms the LBA algorithm proposed in \cite{Junyuan_Beam}, indicating that our proposed data-driven framework of resource allocation outperforms conventional techniques.

\section{Research Challenges and Open Issues}
Machine learning offers a plethora of opportunities for the research in resource allocation for future wireless communications. There are many open issues still not being studied, and need to be further explored. This section outlines some of the most important ones from our viewpoints.

%\subsection{Imbalanced Data}
%
%Canonical machine learning algorithms usually presume that the collected data are distributed uniformly among all classes. However, in many real-life situations, the distribution of data is skewed since representatives of some of classes appear much more frequently, which is called as imbalanced data~\cite{ImbalancedData}. The imbalanced data usually makes the built predictive model to be biased towards the majority class. Therefore, it is more likely to label a new data as the majority class and cause too many classification errors. So far, two methods have been proposed to solve this issue. The first is the data level method which modifies the collection of data to balance distributions in dataset. The second is the algorithm level method which directly modifies existing learning algorithms to alleviate the bias towards majority classes and adapt them to exploit data with skewed distributions. How to use these two methods in our proposed framework remains unexplored.

\subsection{Low-Complexity Classifier}
\label{LowComplexity}
%For the aforementioned $k$-NN algorithm, the simplest method to find nearest neighbors is exhaustive search. If the order of feature vectors is $p$ and the size of training set is $W$, the computational complexity of exhaustive search is $O(pW)$. Some modified $k$-NN algorithms reduce the complexity of searches by restructuring databases into a grid. The grid structuring algorithms can reduce the expected search complexity to $O(log W)$. However, with the increasing of the size of training set, the search complexity will become a huge burden gradually for practical systems. More complicated learning algorithms are required to design lower complexity multiclass classifiers. One of promising algorithms is the support vector machine (SVM) which determines the class of a new feature vector by the use of linear boundaries (hyperplanes) in high dimensional
%spaces~\cite{BIB13}.
%\textcolor[rgb]{1.00,0.00,0.00}{Some approaches about deep learning structures have been provided in literature.
%In \cite{7792369}, a deep neural network structure is proposed.
%In \cite{8088549}, a deep convolutional neural network structure is exploited.
%Although great progress has been made in deep learning, how to reduce the algorithm complexity is a major concern in the future.
%Therefore,}

More advanced techniques are required to design low-complexity multiclass classifiers. One of the promising techniques is to transform the multiclass classification problem into a set of binary classification problems that are efficiently solved using binary classifiers. So far, the support vector machine (SVM) has been regarded as one of the most robust and successful algorithms to design low-complexity binary classifiers, which determines the class of a new feature vector by using linear boundaries (hyperplanes) in high dimensional spaces. More specifically, the two classes are divided by only a few hyperplanes. Accordingly, the class is determined based on which sides of hyperplanes the new feature vector falls into. Compared with the $k$-NN algorithm, the complexity of SVM-based binary classifiers is very low. For the aforementioned beam allocation example, the total number of active beam solutions is $2^N$. In other words, there exist at most $2^N$ classes, which indicates the complexity is $O(2^N)$ for determining the class of scenario. Meanwhile, the complexity of exhaustive search is $O(N^K)$. Obviously, our proposed machine learning framework of resource allocation can approach the optimal performance of exhaustive search with a low complexity. It is worth mentioning that several typical scenarios have been defined with different QoS requirements for future fifth-generation (5G) communications~\cite{Scenarios5G}. For example, ``Great service in a crowd'' scenario focuses on providing reasonable experiences even in crowded areas including stadiums, concerts, and shopping malls. For each typical scenario, the hidden common features on user behaviors and wireless propagation environments may reduce the number of classes, which can be exploited to further reduce the complexity of classifiers. {Recently the deep learning has shown the significant advantages in exploiting the hidden common features~\cite{7792369,8088549}.}

\subsection{Multi-BS Cooperation}
In practical networks, there may exist some users located at the edge of the BS coverage. If the edge user is served by only one BS, the signal quality may be very poor due to the long transmission distance. The cooperative transmission among multiple nearby BSs have been shown to be able to improve the edge user's performance significantly~\cite{CCCRN14}. In this case, the radio resources at multiple BSs should be allocated jointly. Compared with a single BS scenario, the resource allocation problem of multi-BS cooperative transmissions requires more information among multiple BSs. Accordingly, more attributes in historical data will be selected into feature vectors, which makes the learning process more complicated and challenging. How to use historical data to improve the resource allocation with multi-BS cooperative transmissions is very challenging and needs to be studied.

\subsection{Fast Evolution of Scenarios}

In many real scenarios, the user behaviors and wireless environments are time evolving essentially~\cite{OnlineLearning}. That is, the characteristic hidden in historical scenarios is also dynamic. In most cases, such evolutions are too slow and gentle to be noticed. Such slow evolutions can be traced easily by constantly collecting data and periodically updating the dataset for learning. However, in some special situations, the evolutions may be very sudden and significant. For example, an emergency maintenance is carried out for a very busy road which changes the distribution of user locations and mobility characteristics greatly; a high building is demolished by blasting which changes the propagation environments significantly. Since the predictive model is built with outdated historical data, such fast evolutions may result in significant performance loss of resource allocation. In machine learning, this issue can be addressed by updating the predictive model whenever a new data is available. However, since the cloud computing is shared by many applications, the new data can only be stored temporarily at BSs and forwarded to update dataset later. How to deal with the fast evolutions of scenarios in resource allocation is a challenging topic in future research.

\section{Conclusion}
\label{section5}

In future wireless communications, the conventional methods of resource allocation are facing great challenges to meet the ever-increasing QoS requirements of users with scarce radio resource. Inspired by the victory of AlphaGo, this paper proposed a machine learning framework for resource allocation and discussed how to apply the supervised learning to extract the similarities hidden in a great amount of historical data on scenarios. By exploiting the extracted similarities, the optimal or near-optimal solution of the most similar historical scenario is adopted to allocate the radio resources for the current scenario. An example of beam allocation in multi-user massive MIMO systems was then presented to verify that our proposed machine-learning based resource allocation performs better than conventional methods. In a nutshell, machine-learning based resource allocation is an exciting area for future wireless communications assisted by cloud computing.

%\section*{Acknowledgements}
%This work is supported by National Nature Science Foundation of China (Nos. 61172077 \& 61223001), National 863 High Technology Development Project (No. 2013AA013601), Key Special Project of National Science and Technology (No. 2013ZX03003006), and Research Fund of National Mobile Communications Research Laboratory, Southeast University (No. 2013A04).
%

\bibliographystyle{IEEEtran}
\bibliography{MyBibliography}

% that's all folks
\end{document}